\documentclass[conference]{IEEEtran}
\IEEEoverridecommandlockouts
\usepackage[T1]{fontenc}
\usepackage[utf8]{inputenc}
\usepackage{cite}
\usepackage{amsmath,amssymb,amsfonts}
\usepackage{graphicx}
\usepackage{textcomp}
\usepackage{xcolor}
\usepackage{booktabs}
\usepackage{array}
\usepackage{float}
\usepackage{afterpage}
\usepackage[hidelinks]{hyperref}

\definecolor{YaldaBlue}{HTML}{2563EB}
\definecolor{YaldaLightBlue}{HTML}{EFF6FF}
\definecolor{YaldaBorder}{HTML}{93C5FD}
\definecolor{YaldaPurple}{HTML}{5B21B6}
\definecolor{TextGray}{HTML}{334155}
\newcommand{\modelname}{\textsc{Yalda-Embedding}}
\newcommand{\code}[1]{\texttt{#1}}

\newcommand{\traininglossfigure}{%
  \begin{figure*}[!t]
    \centering
    \begin{minipage}[t]{0.32\textwidth}
      \centering
      \includegraphics[width=\linewidth]{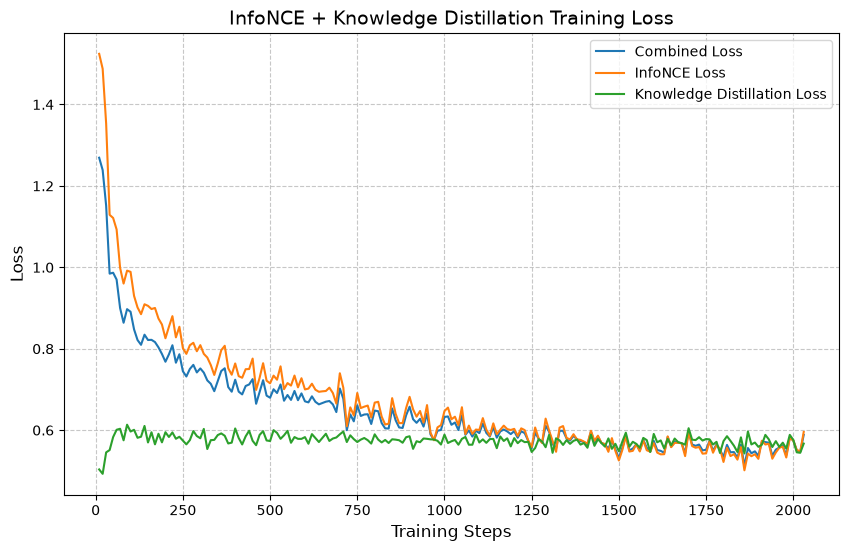}
      \par\smallskip
      {\small (a) Ablation 4: $\alpha=0.25$}
    \end{minipage}\hfill
    \begin{minipage}[t]{0.32\textwidth}
      \centering
      \includegraphics[width=\linewidth]{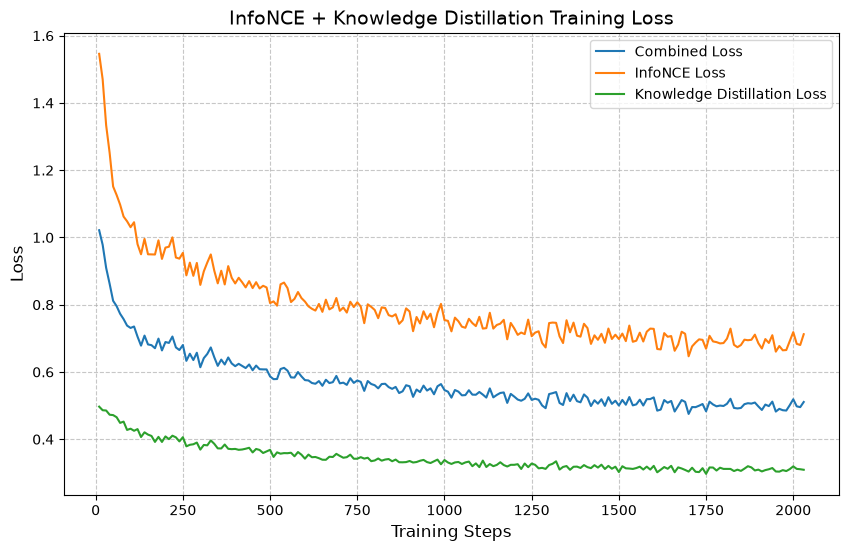}
      \par\smallskip
      {\small (b) Ablation 5: $\alpha=0.50$}
    \end{minipage}\hfill
    \begin{minipage}[t]{0.32\textwidth}
      \centering
      \includegraphics[width=\linewidth]{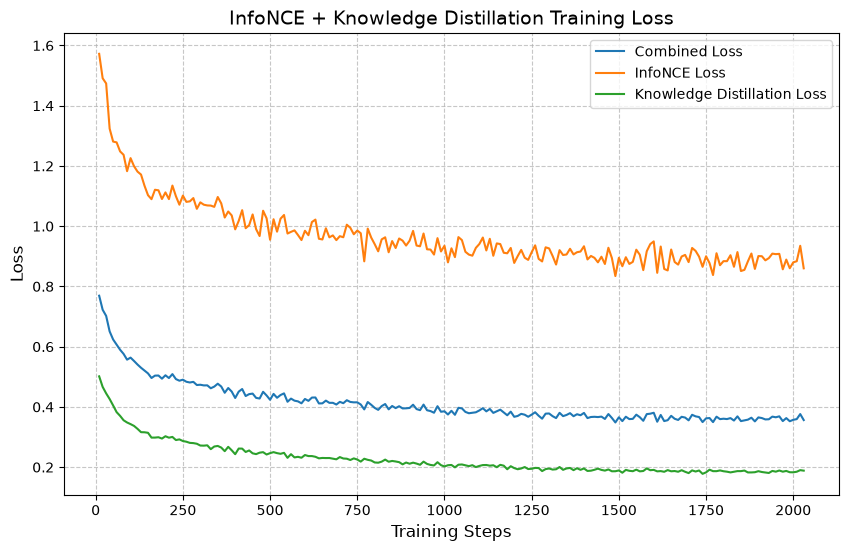}
      \par\smallskip
      {\small (c) Ablation 6: $\alpha=0.75$}
    \end{minipage}
    \caption{Training losses for three knowledge-distillation weights.
    Blue denotes the combined objective, orange the hard-label InfoNCE loss,
    and green the knowledge-distillation loss. Each run uses
    $\mathcal{L}=(1-\alpha)\mathcal{L}_{\mathrm{NCE}}
    +\alpha\mathcal{L}_{\mathrm{KD}}$.}
    \label{fig:training-losses}
  \end{figure*}%
}

\title{Yalda-Embedding: Open Development of a Competitive Persian Retrieval Model}
\author{
\IEEEauthorblockN{Yalda Taheri}
\IEEEauthorblockA{\textit{Faculty of Engineering}\\
\textit{Azad University}\\
y.jaliltaheri@iau.ir}
\and
\IEEEauthorblockN{Mohammad Hassan Heydari}
\IEEEauthorblockA{\textit{Faculty of Computer Engineering}\\
\textit{University of Isfahan}\\
m.heydari@mehr.ui.ac.ir}
\and
\IEEEauthorblockN{Armon Rasooli}
\IEEEauthorblockA{\textit{Department of Electrical Engineering}\\
\textit{Iran University of Science and Technology}\\
a.rasouli@ec.iut.ac.ir}
}

\begin{document}
\maketitle
\afterpage{\afterpage{\traininglossfigure}}

\begin{figure*}[!t]
  \centering
  \includegraphics[width=\textwidth,height=0.34\textheight,keepaspectratio]
    {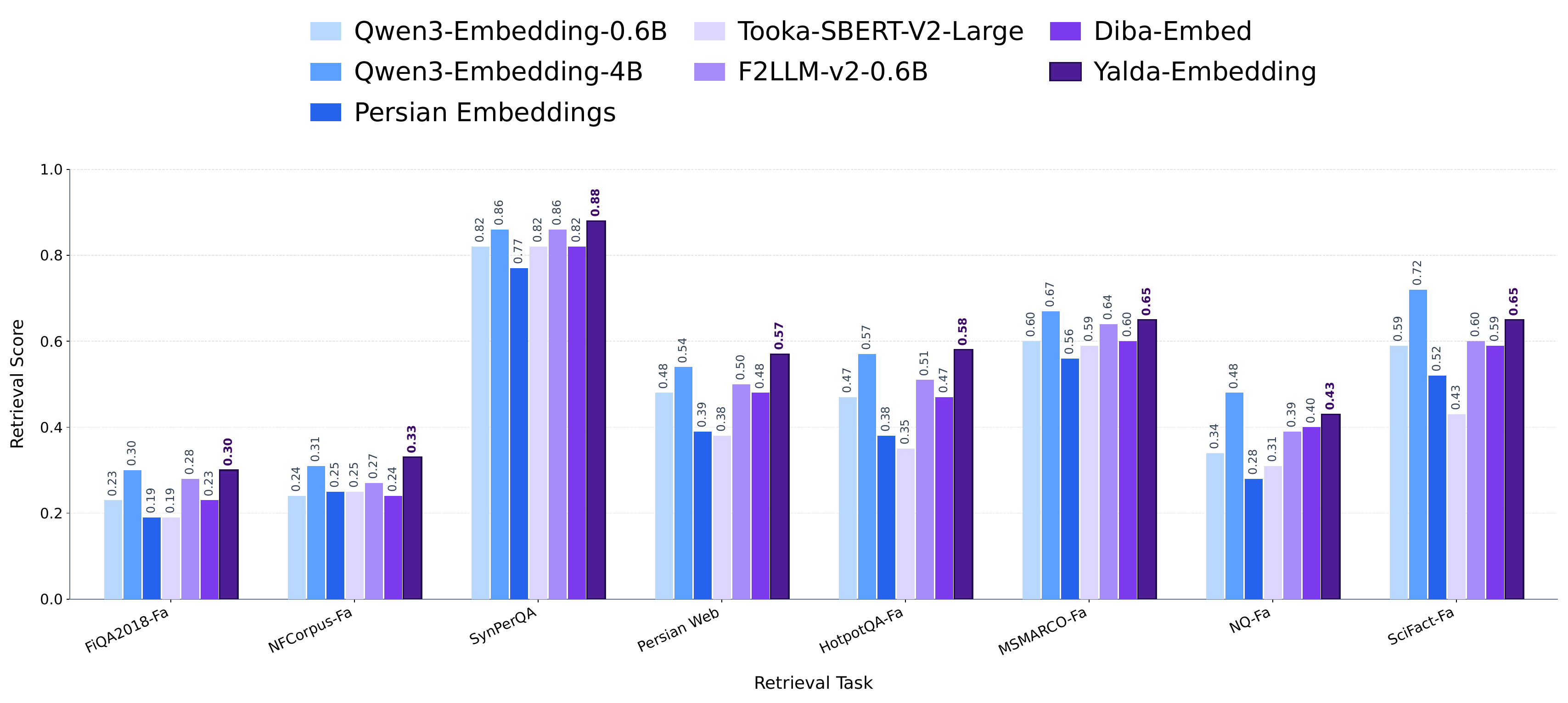}
  \caption{Persian retrieval performance on eight FaMTEB tasks.
  \modelname{} outperforms both same-scale decoder-based baselines on every
  task. Qwen3-Embedding-4B is a higher-capacity reference: \modelname{} wins
  on four tasks and ties on one at the reported precision.}
  \label{fig:main-results}
\end{figure*}

\begin{abstract}
Despite advances in multilingual text representations, effective dense
retrieval for Persian remains constrained by the limited availability of
language-specific training resources and compact embedding models. We present
\modelname, a 595.8-million-parameter Persian retrieval model adapted from
Qwen3-Embedding-0.6B using more than one million query--document pairs.
On eight Persian retrieval tasks from FaMTEB, the model achieves the highest
macro-average among the evaluated 0.6B models, scoring 0.5488 and
outperforming Qwen3-Embedding-0.6B and F2LLM-v2-0.6B on every task. It also
approaches Qwen3-Embedding-4B, winning on four tasks, tying on one, and
trailing by 0.0075 in macro-average. An incremental ablation study examines
contrastive batch size, optimization stabilization, and relational knowledge
distillation from F2LLM-v2-8B. Under the one-epoch training setup, contrastive
fine-tuning alone yields a modest mean improvement over initialization and
regressions on two tasks. The final distilled configuration improves all
eight tasks over the non-distilled model, raising the mean from 0.4813 to
0.5488 without increasing inference-time model size. These gains include
hard-negative evaluation tasks despite training without explicitly mined or
supplied hard negatives. We release model weights, training and evaluation
code, the pairwise dataset, and its teacher-embedded counterpart, enabling
inspection of data composition and training procedures, independent
evaluation, and reuse without repeating expensive teacher inference.
\end{abstract}


\section{Introduction}

Dense text-embedding models map queries and documents to a shared vector space
in which semantic relevance can be estimated efficiently using cosine
similarity or an inner product \cite{reimers2019sentencebert}. They are a central component of semantic
search, retrieval-augmented generation, clustering, recommendation, and
document discovery systems \cite{muennighoff2022mteb,zinvandi2025famteb}.
Although multilingual embedding models have
improved substantially, Persian remains comparatively underrepresented in
both model development and large-scale retrieval training data
\cite{wang2024multilinguale5,zinvandi2025famteb}.

Persian retrieval also faces resource constraints: gathering high-quality
Persian data is challenging, and language-specific training and evaluation
resources remain limited \cite{zinvandi2025famteb}. These
conditions make the adaptation of a multilingual foundation model attractive:
the pretrained model supplies broad semantic and cross-lingual knowledge
\cite{wang2024multilinguale5},
while targeted Persian contrastive training reshapes its representation space
for local retrieval tasks.

This work presents \modelname, an openly developed Persian dense retrieval
model initialized from Qwen3-Embedding-0.6B \cite{zhang2025qwen3embedding}.
The model outperforms both evaluated same-scale decoder-based baselines on
all eight FaMTEB retrieval tasks and approaches the substantially larger
Qwen3-Embedding-4B reference.

The experiments examine how established training choices accumulate under
a fixed student architecture and one epoch of exposure to the training
collection. We first increase the
contrastive batch size from 64 to 512, then stabilize optimization through
gradient clipping, learning-rate warmup, and cosine decay, and finally add
relational knowledge distillation from the much larger F2LLM-v2-8B teacher
\cite{zhang2026f2llmv2}.

The main contributions of this work are:

\begin{itemize}
  \item a 595.8M-parameter Persian retrieval model with the highest mean
  among the evaluated 0.6B models on eight FaMTEB tasks, outperforming both
  same-scale decoder-based baselines on every task and approaching the
  much larger Qwen3-Embedding-4B reference;
  \item a curated training collection of 1,035,981 Persian query--document
  pairs, distributed with precomputed teacher representations;
  \item a two-stage incremental ablation documenting contrastive training
  choices and distillation weights, including the improvement from 0.4813
  without distillation to 0.5488 with the final distilled configuration; and
  \item an open-development release comprising model weights, training and
  evaluation code, and datasets: the code at
  \url{https://github.com/yaldat/yalda-embedding}, the model at
  \url{https://huggingface.co/Yaldat/Yalda-Embedding}, the teacher-embedded
  dataset at
  \url{https://huggingface.co/datasets/Yaldat/Persian-Pairwise-Embedded},
  and the original pairwise dataset at
  \url{https://huggingface.co/datasets/Yaldat/Persian-Pairwise}.
\end{itemize}

These assets make the development process inspectable: readers can examine
the source mixture and query--document pairs, reuse precomputed teacher
representations, inspect the optimization and benchmarking procedures, and
evaluate the released weights. The teacher-enriched dataset also allows
subsequent student experiments without repeating the offline 8B teacher
inference stage.

\section{Related Work}

\subsection{Sentence and Text Embeddings}

Sentence-BERT established the modern bi-encoder formulation for independently
encoding texts and comparing their fixed-size representations
\cite{reimers2019sentencebert}. SimCSE demonstrated that contrastive learning
can substantially improve sentence representations using either dropout-based
views or supervised natural-language-inference pairs \cite{gao2021simcse}.
E5 later scaled weakly supervised contrastive pretraining across diverse text
pairs \cite{wang2022e5}, with multilingual E5 extending this recipe across
languages and retrieval settings \cite{wang2024multilinguale5}.

General-purpose embedding systems increasingly combine heterogeneous tasks,
instructions, and multi-stage contrastive learning. GTE uses large and diverse
training mixtures in a staged contrastive pipeline \cite{li2023gte}, while
synthetic instruction generation has enabled strong embedding models with
relatively short supervised fine-tuning schedules
\cite{wang2024improving}. MTEB provides a unified evaluation framework across
embedding tasks \cite{muennighoff2022mteb}; FaMTEB extends this effort to
Persian and introduces a broad set of Persian evaluation resources
\cite{zinvandi2025famteb}. Tooka-SBERT is a lightweight Persian sentence
embedding family based on the TookaBERT encoder and provides a directly
relevant Persian baseline \cite{zamaninejad2025tookasbert}.

\subsection{Decoder-Only Models as Embedding Encoders}

Decoder-only language models can be adapted into effective embedding models.
SGPT investigated pooling and contrastive adaptation of GPT-style models for
semantic search \cite{muennighoff2022sgpt}. LLM2Vec studied bidirectional
attention, masked next-token prediction, and contrastive learning for turning
decoder-only language models into text encoders
\cite{behnamghader2024llm2vec}. The Qwen3 Embedding family instead adapts the
Qwen3 foundation models through large-scale pretraining, supervised
fine-tuning, synthetic multilingual data, and model merging
\cite{zhang2025qwen3embedding,yang2025qwen3}.

The student used here retains the Qwen3-Embedding decoder architecture. Its
components follow the Transformer design \cite{vaswani2017attention} and
include rotary position embeddings \cite{su2021roformer}, grouped-query
attention \cite{ainslie2023gqa}, SwiGLU feed-forward layers
\cite{shazeer2020glu}, and RMS normalization \cite{zhang2019rmsnorm}.

\subsection{Contrastive Learning and Knowledge Distillation}

InfoNCE was introduced in contrastive predictive coding as a probabilistic
objective for distinguishing related representations from sampled negatives
\cite{oord2018cpc}. In dense retrieval, a minibatch naturally supplies many
negative documents for each query, making larger batches especially valuable.
Related supervised contrastive objectives likewise show the benefit of
learning against multiple positives and negatives in a shared batch
\cite{khosla2020supervised}.

Knowledge distillation transfers behavior from a larger teacher into a smaller
student, commonly by matching softened output distributions
\cite{hinton2015distilling}. Retrieval-specific distillation can transfer
relevance behavior while preserving independent document encodings and fast
vector search \cite{shakeri2019kdretrieval}. F2LLM-v2 combines multilingual
embedding training with Matryoshka representations and knowledge distillation
across a family of model sizes \cite{zhang2026f2llmv2}. Matryoshka
representation learning organizes information into useful nested prefixes,
allowing shorter embedding dimensions to retain semantic utility
\cite{kusupati2022matryoshka}.

\section{Dataset}

\subsection{Persian Pairwise Collection}

We assembled 1,035,981 Persian query--positive-document pairs from 13 retrieval
resources. We normalized every source to a common pairwise schema with a
\code{question} field and a \code{context} field and globally shuffled the
combined collection before training. Table~\ref{tab:data-mixture} reports the composition
of the final collection.

\begin{table}[!t]
  \centering
  \caption{Composition of the Persian pairwise training collection.}
  \label{tab:data-mixture}
  \small
  \resizebox{\columnwidth}{!}{%
\begin{tabular}{lrr}
    \toprule
    Source & Pairs & Share (\%) \\
    \midrule
    MSMARCO-Fa & 532,501 & 51.40 \\
    HotpotQA-Fa & 169,946 & 16.40 \\
    NFCorpus-Fa & 110,575 & 10.67 \\
    PersianWebDocumentRetrieval & 78,871 & 7.61 \\
    PQuAD & 63,994 & 6.18 \\
    Synthetic Persian QA Retrieval & 27,243 & 2.63 \\
    PersianQuAD & 18,567 & 1.79 \\
    FiQA2018-Fa & 14,130 & 1.36 \\
    ArguAna-Fa v2 & 8,674 & 0.84 \\
    ClimateFEVER-Fa & 4,679 & 0.45 \\
    NQ-Fa & 4,201 & 0.41 \\
    CQADupStack-Unix-Fa & 1,681 & 0.16 \\
    SciFact-Fa & 919 & 0.09 \\
    \midrule
    \textbf{Total} & \textbf{1,035,981} & \textbf{100.00} \\
    \bottomrule
  \end{tabular}
}
\end{table}

The source mixture is intentionally not resampled: the model observes the
empirical union once. Consequently, large web-retrieval collections dominate
the update count, while smaller scientific and financial collections provide
domain diversity. The large global batch also means that every query uses the
other 511 documents in its batch as negatives. This is computationally
efficient, but it can introduce false negatives when two documents happen to
answer semantically related queries.

\subsection{Teacher-Embedded Release}

We release the teacher-embedded version of the pairwise data on Hugging Face at
\url{https://huggingface.co/datasets/Yaldat/Persian-Pairwise-Embedded}.
For each query and document, we precomputed an F2LLM-v2-8B representation. The
teacher originally produces a 4,096-dimensional Matryoshka embedding; we
retain its first 1,024 dimensions and renormalize the result before
constructing teacher similarity scores.

Embedding the complete collection required approximately 250 hours on a
single NVIDIA P100 GPU and produced roughly 17.5~GB of teacher-enriched
training data across 36 Parquet shards. Performing this expensive teacher
inference once, offline, avoids loading the 8B teacher during TPU training and
makes subsequent student experiments substantially cheaper.

We assembled Persian-Pairwise exclusively from the training splits of its
constituent sub-datasets. We perform all evaluations reported in this work
on the test splits of the corresponding evaluation datasets. Training and evaluation
therefore use separate source-provided partitions.

\section{Methodology}

\subsection{Student Architecture}

\modelname{} is initialized from Qwen3-Embedding-0.6B. Table~\ref{tab:model}
summarizes the student architecture implemented for training.

\begin{table}[!t]
  \centering
  \caption{Student architecture.}
  \label{tab:model}
  \resizebox{\columnwidth}{!}{%
\begin{tabular}{lr}
    \toprule
    Property & Value \\
    \midrule
    Parameters & 595,776,512 \\
    Transformer blocks & 28 \\
    Vocabulary size & 151,669 \\
    Hidden / embedding size & 1,024 \\
    Feed-forward intermediate size & 3,072 \\
    Query attention heads & 16 \\
    Key/value attention heads & 8 \\
    Attention head dimension & 128 \\
    Positional encoding & RoPE \\
    RoPE base & 1,000,000 \\
    Attention pattern & Causal \\
    Pooling & Last non-padding token \\
    Maximum training length & 512 tokens \\
    \bottomrule
  \end{tabular}
}
\end{table}

Queries are formatted with the fixed instruction

\begin{quote}
\small\ttfamily
Instruct: Given a web search query, retrieve relevant passages that answer the
query\\
Query:<query>
\end{quote}

while documents are encoded without an instruction. Inputs are left-padded and
truncated to 512 tokens. The final hidden state at the last valid token is used
as the text representation. The language-modeling head is omitted.

\subsection{InfoNCE Objective}

For a minibatch of $B$ matched query--document pairs
$\{(q_i,d_i)\}_{i=1}^{B}$, the shared student encoder $f_{\theta}$ produces
embeddings that are normalized to unit length:

\begin{equation}
  \hat{q}_i = \frac{f_{\theta}(q_i)}{\lVert f_{\theta}(q_i) \rVert_2},
  \qquad
  \hat{d}_j = \frac{f_{\theta}(d_j)}{\lVert f_{\theta}(d_j) \rVert_2}.
\end{equation}

Student query--document logits are cosine similarities scaled by the student
temperature $\tau_s=0.05$:

\begin{equation}
  s_{ij} = \frac{\hat{q}_i^{\mathsf{T}}\hat{d}_j}{\tau_s}.
\end{equation}

The positive document for query $i$ is at diagonal position $i$; all other
documents in the batch are treated as negatives. The hard-label InfoNCE loss is

\begin{equation}
  \mathcal{L}_{\mathrm{NCE}}
  = -\frac{1}{B}\sum_{i=1}^{B}
  \log\frac{\exp(s_{ii})}{\sum_{j=1}^{B}\exp(s_{ij})}.
  \label{eq:infonce}
\end{equation}

The implementation uses only the query-to-document direction rather than
adding a symmetric document-to-query term. Training uses query--positive-document
pairs and in-batch negatives only; no explicitly mined or supplied hard
negatives are used. The hard-label objective refers to the diagonal positive
labels, not to hard-negative sampling.

\subsection{Relational Knowledge Distillation}

Let $g(q_i)$ and $g(d_j)$ denote the offline teacher embeddings. After taking
the first 1,024 Matryoshka dimensions, the teacher representations are
independently L2-normalized. Teacher logits use temperature $\tau_t=0.07$:

\begin{equation}
  t_{ij} =
  \frac{\widehat{g(q_i)}^{\mathsf{T}}\widehat{g(d_j)}}{\tau_t}.
\end{equation}

For every query, the teacher and student define distributions over all
documents in the batch:

\begin{equation}
  p^T_{ij}=\operatorname{softmax}_{j}(t_{ij}),
  \qquad
  p^S_{ij}=\operatorname{softmax}_{j}(s_{ij}).
\end{equation}

We minimize the forward Kullback--Leibler divergence from the teacher
distribution to the student distribution:

\begin{equation}
  \mathcal{L}_{\mathrm{KD}}
  = \frac{1}{B}\sum_{i=1}^{B}
  \sum_{j=1}^{B}p^T_{ij}
  \left(\log p^T_{ij}-\log p^S_{ij}\right).
  \label{eq:kd}
\end{equation}

The final objective uses $\alpha=0.75$; the earlier Ablations~4 and~5 use
$\alpha=0.25$ and $\alpha=0.50$, respectively, and are retained in the
ablation study:

\begin{equation}
  \mathcal{L}
  = (1-\alpha)\mathcal{L}_{\mathrm{NCE}}
  + \alpha\mathcal{L}_{\mathrm{KD}}
  = 0.25\mathcal{L}_{\mathrm{NCE}}
  + 0.75\mathcal{L}_{\mathrm{KD}}.
  \label{eq:combined}
\end{equation}

This formulation is relational: it transfers the teacher's ranking geometry
within each batch instead of forcing the student to regress directly onto
teacher coordinates. It can therefore be used even when the teacher and
student have different native embedding dimensions. The 1,024-dimensional
teacher prefix is retained here primarily to reduce data-transfer and storage
costs and to align the deployed embedding width.

\section{Training}

\subsection{Optimization}

The complete pretrained student is fine-tuned for one epoch using AdamW
\cite{loshchilov2017adamw}. The peak learning rate is $2\times10^{-5}$. A
linear warmup covers the first 10\% of updates, followed by cosine decay to
zero for the remainder of training. Gradients are clipped to a global norm of
1.0. The global batch size is 512, producing a $512\times512$ similarity matrix
for both student and teacher on every step.

The dataset contains 1,035,981 examples. Using complete batches results in
2,023 optimization steps and 1,035,776 consumed examples; 205 examples from
the final incomplete batch are omitted. The data are reshuffled through a
NumPy permutation. The model random seed is fixed to zero, but the dataset
permutation seed is not fixed in the final run.

\begin{table}[!t]
  \centering
  \caption{Final training configuration.}
  \label{tab:training}
  \resizebox{\columnwidth}{!}{%
\begin{tabular}{lr}
    \toprule
    Hyperparameter & Value \\
    \midrule
    Epochs & 1 \\
    Global batch size & 512 \\
    Batch size per TPU device & 64 \\
    Optimization steps & 2,023 \\
    Peak learning rate & $2\times10^{-5}$ \\
    Warmup & 10\% of updates \\
    Decay & Cosine to zero \\
    Gradient clipping & Global norm 1.0 \\
    Student temperature & 0.05 \\
    Teacher temperature & 0.07 \\
    KD weight $\alpha$ & 0.75 \\
    Maximum sequence length & 512 \\
    \bottomrule
  \end{tabular}
}
\end{table}

\subsection{Mixed Precision}

Training follows a mixed-precision policy \cite{micikevicius2017mixed}.
Model parameters, gradients, AdamW first and second moments, normalization,
rotary-position calculations, attention softmax, embedding normalization, and
loss calculations use FP32. Memory-intensive, numerically less-sensitive
forward activations and matrix multiplications use BF16. This arrangement
reduces activation memory and takes advantage of TPU BF16 throughput while
retaining FP32 precision for the state and operations most vulnerable to
numerical instability.

\subsection{Distributed Execution}

The final run uses eight TPU v5e devices, each with 16~GB of high-bandwidth
memory. JAX creates a one-dimensional device mesh with a single sharding axis.
The minibatch is partitioned across this axis, placing 64 query--document pairs
on each device. Large parameter matrices are also partitioned along selected
input or output dimensions, while small normalization parameters are
replicated. This constitutes single-axis fully sharded data parallelism
(FSDP): parameters, optimizer state, and corresponding updates are distributed
across all eight devices.

The model and training step are implemented in JAX 0.9.2 and Flax NNX 0.12.6.
The complete update is compiled with \code{nnx.jit}. Every Transformer block is
wrapped with activation rematerialization, trading additional computation for
lower activation memory. CPU tokenization and loading are overlapped with TPU
execution using a background prefetch queue of ten batches. The recorded
Ablation~4 training run ($\alpha=0.25$) took approximately 5.06 hours, with a steady-state median of about
8.55 seconds per optimization step after initial compilation.

All ablations and training runs together required approximately 60 hours
of execution on eight TPU v5e devices, equivalent to 480 TPU v5e
device-hours. This total includes the approximately five-hour training run
of the final model; it is not an additional cost beyond that run. The
offline teacher-embedding computation on the NVIDIA P100 is reported
separately in the Dataset section.

\subsection{Checkpoint Export and Availability}

After training, Flax kernels are transposed into the PyTorch tensor layout and
saved as FP32 Safetensors. The tokenizer, configuration, and supporting files
are inherited from the original Qwen3-Embedding-0.6B repository. The model is
available at \url{https://huggingface.co/Yaldat/Yalda-Embedding}.

\section{Results}

\subsection{Evaluation Setup}

We evaluate on eight Persian retrieval tasks from the FaMTEB benchmark
\cite{zinvandi2025famteb}: FiQA2018-Fa, NFCorpus-Fa,
SynPerQARetrieval, PersianWebDocumentRetrieval, HotpotQA-FaHardNegatives,
MSMARCO-FaHardNegatives, NQ-FaHardNegatives, and SciFact-Fa. We use
the test splits and load models through Sentence Transformers in FP16 on a
GPU. We set a maximum input length of 512 tokens and use an encoding batch
size of 64. We retain each model's prompting and pooling behavior as defined
by its configuration and MTEB integration, rather than imposing a common
query instruction across models. We use MTEB for retrieval scoring and
report the retrieval scores returned by the MTEB library
for the selected FaMTEB tasks. The summary is their unweighted macro-average,
not a score over the full FaMTEB suite.

We record dependency versions during evaluation and install MTEB without a
version pin. Table~\ref{tab:baseline-averages} reports the MTEB-reported retrieval
scores at their available precision; the comparisons do not establish
statistical significance.

The selected baselines cover the student initialization, a higher-capacity
Qwen reference, Persian-specific encoder models, and the same-scale F2LLM-v2
student family. We also evaluate \code{Dibachain/Diba-Embed} as an additional
baseline. Figure~\ref{fig:main-results} presents per-task performance;
Table~\ref{tab:baseline-averages} summarizes macro-averages.

\begin{table}[!t]
  \centering
  \caption{Macro-average on eight FaMTEB retrieval tasks, grouped by
  comparison role. Higher is better. Sizes are nominal parameter counts.
  Bold marks the best same-scale result.}
  \label{tab:baseline-averages}
  \resizebox{\columnwidth}{!}{%
\begin{tabular}{lrr}
    \toprule
    Model & Size & Mean \\
    \midrule
    \multicolumn{3}{l}{\textit{Same-scale comparison}} \\
    \modelname{} & 0.6B & \textbf{0.5488} \\
    F2LLM-v2-0.6B & 0.6B & 0.5063 \\
    Dibachain/Diba-Embed & 0.6B & 0.4788 \\
    Qwen3-Embedding-0.6B & 0.6B & 0.4713 \\
    heydariAI/persian-embeddings & 0.6B & 0.4175 \\
    \midrule
    \multicolumn{3}{l}{\textit{Smaller Persian baseline}} \\
    Tooka-SBERT-V2-Large & 0.4B & 0.4150 \\
    \midrule
    \multicolumn{3}{l}{\textit{Higher-capacity reference}} \\
    Qwen3-Embedding-4B & 4B & 0.5563 \\
    \bottomrule
  \end{tabular}
}
\end{table}

\modelname{} achieves the highest mean among the evaluated 0.6B models,
outperforming Qwen3-Embedding-0.6B, F2LLM-v2-0.6B,
\code{Dibachain/Diba-Embed}, and \code{heydariAI/persian-embeddings} on
all eight tasks.
It improves the Qwen3-Embedding-0.6B initialization by 0.0775
absolute, or approximately 16.4\% relative, and performs better on all eight
tasks. The final model also exceeds F2LLM-v2-0.6B by 0.0425 in macro-average
and outperforms it on all eight tasks at the reported precision.

Compared with \code{Dibachain/Diba-Embed}, which obtains a macro-average of
0.4788, \modelname{} improves the mean by 0.0700 absolute (approximately
14.6\% relative) and achieves higher scores on all eight tasks. The largest
per-task gains are on HotpotQA-FaHardNegatives ($+0.11$), NFCorpus-Fa
($+0.09$), and PersianWebDocumentRetrieval ($+0.09$). On the three
hard-negative tasks, the gains over Diba-Embed are $+0.11$ for HotpotQA,
$+0.05$ for MSMARCO, and $+0.03$ for NQ.

Against the higher-capacity Qwen3-Embedding-4B reference (mean 0.5563),
\modelname{} wins on four tasks, ties on one, and trails on three.
Despite being substantially smaller, \modelname{} exceeds it on
NFCorpus-Fa, SynPerQARetrieval, PersianWebDocumentRetrieval, and
HotpotQA-FaHardNegatives, and ties it on FiQA2018-Fa. Its macro-average is
0.0075 below the 4B reference.

Notably, these results are obtained without explicit hard-negative training.
On HotpotQA-FaHardNegatives, MSMARCO-FaHardNegatives, and NQ-FaHardNegatives,
\modelname{} scores 0.58, 0.65, and 0.43, respectively, outperforming both
evaluated 0.6B decoder-based baselines on all three tasks. This demonstrates
competitive performance on hard-negative evaluation tasks using only in-batch
negatives and relational distillation during student training.

\subsection{Ablation Study}

The model was developed through six cumulative configurations in two stages:
contrastive training recipe development (Ablations~1--3) and a
distillation-weight study (Ablations~4--6). Ablation~1
uses batch size 64, InfoNCE, a fixed learning rate, and no gradient clipping.
Ablation~2 changes only the global batch size to 512. Ablation~3 adds global
gradient clipping, 10\% linear warmup, and cosine decay. Ablation~4 adds
relational distillation from F2LLM-v2-8B with $\alpha=0.25$ while retaining
the Ablation~3 training configuration. Ablation~5 changes only the KD weight
to $\alpha=0.50$. The final model, Ablation~6, increases this weight to
$\alpha=0.75$ while retaining the remaining training configuration.

\begin{table*}[!t]
  \centering
  \caption{Two-stage incremental ablation: contrastive recipe development
  followed by a distillation-weight study. Values are retrieval scores;
  $\Delta$ is the change in macro-average from the previous row.}
  \label{tab:ablation}
  \resizebox{\textwidth}{!}{%
  \begin{tabular}{llrrrrrrrrrr}
    \toprule
    Ablation & Change & FiQA & NFC & Syn & Web & Hotpot & MSM & NQ & SciFact & Mean & $\Delta$ \\
    \midrule
    \multicolumn{12}{l}{\textit{Stage I: Contrastive training recipe (no KD)}} \\
    1 & InfoNCE, BS 64
      & 0.18 & 0.29 & 0.80 & 0.47 & 0.41 & 0.59 & 0.28 & 0.46 & 0.4350 & -- \\
    2 & Batch size 512
      & 0.24 & 0.31 & 0.82 & 0.54 & 0.47 & 0.60 & 0.29 & 0.52 & 0.4738 & +0.0388 \\
    3 & Clip + warmup + cosine
      & 0.25 & 0.31 & 0.83 & 0.54 & 0.48 & 0.61 & 0.32 & 0.51 & 0.4813 & +0.0075 \\
    \midrule
    \multicolumn{12}{l}{\textit{Stage II: Distillation weights (Stage I recipe fixed)}} \\
    4 & Ablation 3 + KD, $\alpha=0.25$
      & 0.26 & 0.33 & 0.86 & 0.56 & 0.55 & 0.62 & 0.35 & 0.57 & 0.5125 & +0.0313 \\
    5 & KD, $\alpha=0.50$
      & 0.28 & 0.32 & 0.87 & 0.57 & 0.57 & 0.64 & 0.39 & 0.61 & 0.5313 & +0.0188 \\
    6 (final) & KD, $\alpha=0.75$
      & 0.30 & 0.33 & 0.88 & 0.57 & 0.58 & 0.65 & 0.43 & 0.65 & \textbf{0.5488} & \textbf{+0.0175} \\
    \bottomrule
  \end{tabular}%
  }
\end{table*}

The central comparison is between the stabilized non-distilled recipe
(Ablation~3) and the final distilled model (Ablation~6). Under the one-epoch
setup, contrastive fine-tuning alone raises the mean from the initialization's
0.4713 to 0.4813, but lowers NQ from 0.34 to 0.32 and SciFact from 0.59 to
0.51. The final distilled configuration raises the mean to 0.5488, a gain of
0.0675 over Ablation~3, and improves all eight tasks over both Ablation~3 and
the initialization. Thus, the observed benefit of distillation extends beyond
a mean increase to reversing these task-specific regressions.

Increasing the batch size produces the largest individual average gain,
$+0.0388$. With InfoNCE, this intervention increases the number of in-batch
negative documents per query from 63 to 511. The stabilization bundle adds a
smaller but consistent mean improvement of $+0.0075$. Since clipping, warmup,
and cosine decay were introduced together, this experiment does not identify
their separate effects.

Relational KD with $\alpha=0.25$ adds $+0.0313$ to the macro-average and improves every evaluated
task relative to Ablation~3. Its largest gains occur on HotpotQA
($+0.07$) and SciFact ($+0.06$), followed by SynPerQA and NQ ($+0.03$ each).
Increasing the KD weight to $\alpha=0.50$ adds a further $+0.0188$ to the
macro-average, improving seven of eight tasks over Ablation~4. NFCorpus-Fa
decreases by 0.01. The largest gains are on NQ and SciFact ($+0.04$ each).
Increasing the KD weight again to $\alpha=0.75$ adds $+0.0175$ to the
macro-average, improving seven tasks over Ablation~5 while leaving
PersianWebDocumentRetrieval unchanged. The largest gains are again on NQ and
SciFact ($+0.04$ each). Across the complete development chain, the mean rises
from 0.4350 to 0.5488, an absolute gain of 0.1138 and a relative gain of
approximately 26.1\%.

\subsection{Training Dynamics}

Figure~\ref{fig:training-losses} shows the combined objective and its InfoNCE
and KD components for Ablations~4--6, with $\alpha=0.25$, $0.50$, and $0.75$,
respectively. All three runs exhibit an initial rapid decrease followed by
more gradual changes. Increasing the KD weight yields a lower teacher--student
divergence but a higher final hard-label InfoNCE loss, illustrating the
trade-off between matching teacher relevance distributions and fitting the
diagonal positive labels. Combined-loss magnitudes should not be compared
directly across runs because their component weights differ.

The following numerical summaries refer only to the recorded Ablation~4 run
with $\alpha=0.25$; the plots for Ablations~5 and~6 provide qualitative
comparisons rather than additional interval-level numerical summaries.
The first ten logged intervals, covering approximately the first 100 updates,
have mean losses of 1.015 for the combined objective, 1.165 for InfoNCE, and
0.567 for KD. Over the final ten intervals, these values are 0.559, 0.557, and
0.564, respectively. Most of the observed decrease therefore comes from the
hard-label retrieval objective. The teacher divergence remains close to its
early value, indicating that the teacher continues to impose a nontrivial soft
ranking constraint throughout training rather than becoming trivially
matched.

\section{Conclusion}

We introduced \modelname, an openly developed Persian retrieval model
adapted from Qwen3-Embedding-0.6B on more than one million query--document
pairs. On eight FaMTEB retrieval tasks, it achieves the highest mean among
the evaluated 0.6B models (0.5488) and outperforms both same-scale
decoder-based baselines on every task. It approaches Qwen3-Embedding-4B,
winning on four tasks and tying on one, with a macro-average gap of 0.0075.

The two-stage ablation documents the progression from contrastive training
to relational distillation. The final distilled model improves every task
over the non-distilled recipe and raises its mean from 0.4813 to 0.5488
while retaining the same inference architecture. By releasing weights,
training and evaluation code, training pairs, and reusable teacher
representations, this work makes both the resulting model and the choices
behind its development available for independent inspection and extension.

\bibliographystyle{IEEEtran}
\bibliography{references}

\end{document}